\pdfoutput=1

\documentclass[11pt]{article}

\usepackage[final]{acl}

\usepackage{times}
\usepackage{latexsym}
\usepackage{graphicx} 
\usepackage{tcolorbox}
\usepackage{amsmath}
\usepackage{multirow}
\usepackage{subcaption} 

\usepackage[ruled,linesnumbered]{algorithm2e}

\usepackage{amsmath}
\usepackage{url}
\usepackage[export]{adjustbox}
\usepackage{float}
\usepackage{amssymb}  
\usepackage[utf8]{inputenc}  
\usepackage{csquotes}

\usepackage{float}
\usepackage[T1]{fontenc}

\usepackage[utf8]{inputenc}

\usepackage{microtype}

\usepackage{inconsolata}
\usepackage{todonotes}

\title{Knowledge-Driven Cross-Document Relation Extraction}

\author{
  Monika Jain \and Raghava Mutharaju \\
  Knowledgeable Computing and Reasoning Lab, IIIT-Delhi, India \\
  \texttt{\{monikaja, raghava.mutharaju\}@iiitd.ac.in} \\
  \textbf{Kuldeep Singh} \\
  Cerence GmbH and Zerotha Research, Germany \\
  \texttt{kuldeep.singh1@cerence.com} \\
  \textbf{Ramakanth Kavuluru} \\
  University of Kentucky, Lexington, Kentucky, United States \\
  \texttt{ramakanth.kavuluru@uky.edu} \\ }

\begin{document}
\maketitle

\begin{abstract}
Relation extraction (RE) is a well-known NLP application often treated as a sentence- or document-level task. However, a handful of recent efforts explore it across documents or in the cross-document setting (CrossDocRE). This is distinct from the single document case because different documents often focus on disparate themes, while text within a document tends to have a single goal. Linking findings from disparate documents to identify new relationships is at the core of the popular literature-based knowledge discovery paradigm in biomedicine and other domains. Current CrossDocRE efforts do not consider domain knowledge, which are often assumed to be known to the reader when documents are authored. Here, we propose a novel approach, KXDocRE, that embed domain knowledge of entities with input text for cross-document RE. Our proposed framework has three main benefits over baselines: 1) it incorporates domain knowledge of entities along with documents' text; 2) it offers interpretability by producing explanatory text for predicted relations between entities  3) it improves performance over the prior methods.
Code and models are available at~\url{https://github.com/kracr/cross-doc-relation-extraction}

\end{abstract}

\section{Introduction}

Identifying relations between entity pairs within unstructured text is known as relation extraction (RE). Earlier studies concentrated on uncovering relations between entities within a sentence~\cite{zhang-etal-2017-position, Hsu2022ASA, zeng-etal-2014-relation,dos-santos-etal-2015-classifying,cai-etal-2016-bidirectional,10.1007/978-3-031-43421-1_14}. However, an entity pair need not necessarily occur in a sentence. It can also be part of a document. There is substantial literature on extracting relations between entities based on the document itself~\cite{info14070365,Xu2020DocumentLevelRE,christopoulou-etal-2019-connecting,nan-etal-2020-reasoning,zeng-etal-2020-double,li-etal-2021-mrn,jain2024revisiting}. Beyond a single sentence or a document, entity pairs can also be present in different documents~\cite{yao-etal-2021-codred}.

\begin{figure}[ht]
    \centering
    \includegraphics[width=0.48\textwidth]{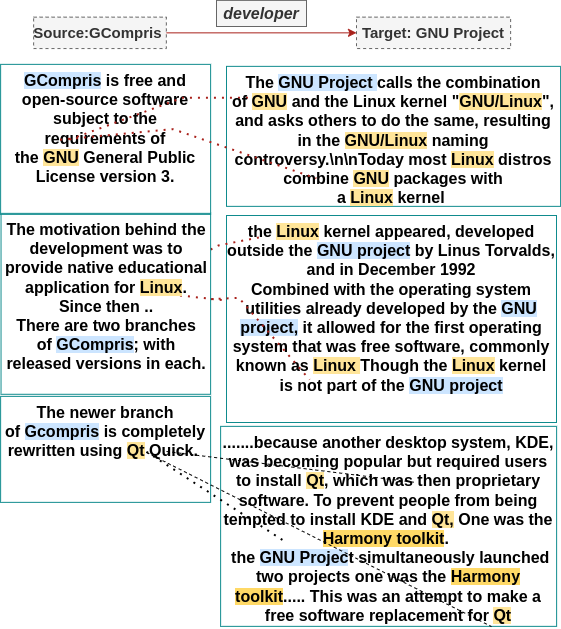}
\caption{Three text paths indicate the relationship path between the source entity, GCompris, and the target entity, GNU Project. These connections are established through pairs of documents, where one document features the source entity, and the other contains the target entity. In each path, the connection between the source and target entities is led by a mentioned entity in both documents (e.g., Linux).}
    \label{fig:mesh1}
\end{figure}

An analysis on Wikipedia shows that more than 57.6 relational facts do not co-occur in a single document~\cite{10.1145/2629489}. This indicates that many facts occur across the documents. Considering that, little work has been performed in cross-document relation extraction (CrossDocRE). CodRED (cross-document relation extraction dataset) is the first dataset published in this line of work~\cite{yao-etal-2021-codred}, which serves as a starting point to solve CrossDocRE. Documents containing the source entity are identified and retrieved from multiple documents; the same is done for the target entity. Various text paths between entity pairs (source and target entity) across documents are recognized using the mentioned entity (other than the source and target entity). Text paths refer to the paths that connect the source and target entities via the mentioned entities. These paths are retrieved from both the source and target documents. 

Figure~\ref{fig:mesh1} shows an example of CrossDocRE. Between entity pairs \textbf{GCompris} and \textbf{GNU Project}, multiple text paths via the mentioned entities, such as \emph{Linux, Qt} and \emph{GNU}, can be used to get the correct relation label as \textbf{developer}. CodRED uses these text paths in a bag and performs reasoning~\cite{yao-etal-2021-codred}. Although this approach seems reasonable, along with relevant text, these text paths also contain noisy data. To overcome these issues, ECRIM (entity-centered cross-document relation extraction) proposed filtering text paths using a mentioned entity~\cite{wang-etal-2022-entity}. However, ECRIM only works in one of the settings of CrossDocRE where text paths are provided for reasoning. To address this issues, MR.COD has been proposed, which is a multi-hop reasoning framework based on path mining and ranking~\cite{DBLP:conf/acl/LuHZMC23}. These models rely on the knowledge between entities in a text and do not consider the domain knowledge associated with entities. 
Past work along these lines uses entity types and entity aliases to predict the relation~\cite{10.1007/978-3-030-62419-4_11}. RECON~\cite{bastos2021recon} encoded attribute and relation triples in the Knowledge Graph. KB-Both~\cite{inproceedings} uses entity details from hyperlinked text documents of Wikipedia and Knowledge Graph (KG) from Wikidata to enhance performance. Given input text as a sentence or a document, these models use basic details of entities to improve the performance. In this paper, we explore whether incorporating domain knowledge can enhance the performance of CrossDocRE tasks. The main contributions of this work are as follows.

\begin{itemize}
    \item We propose a novel model that integrates domain knowledge with a cross-document relation extraction model.
\item  Our validation demonstrates the  effectiveness of our model, with nontrivial  performance gains in cross-document RE. 
\item  We enhance the predicted relationships through textual explanations, offering insights into how the relations were predicted.

\end{itemize}

\section{Related work}

The relation between entities can extend across multiple documents, and researchers have investigated the extraction of entities, events, and relationships in a cross-document context with unlabeled data~\cite{10.5555/1870658.1870757}.
CodRED~\cite{yao-etal-2021-codred} is the first open source human-annotated cross-document dataset. In this work, the authors address the problem using two approaches. The first method involves a pipeline approach in which they construct a relational graph for each document and reason over these graphs to extract the desired relation. The second method, referred as the joint approach, combines various text path representations through a selective attention mechanism to predict relations. Although this method is effective, it does not consider the mentioned entity-based sentences. ECRIM uses an entity-based document context filter to retain useful information in the given documents by using the mentioned entities in the text paths. Secondly, they solve CrossDocRE using cross-path entity relation attention, allowing entity relations across text paths to interact with each other~\cite{wang-etal-2022-entity}. Nevertheless, this work focuses on a closed setting where evidential context has been given instead of all documents. A multi-hop evidence retrieval method based on evidence path mining and ranking has been proposed in MR.COD~\cite{DBLP:conf/acl/LuHZMC23}. In evidence path mining, a multi-document passage graph is constructed, where passages are linked by edges
corresponding to shared entities. A graph traversal algorithm mines the passage paths from source-to-target entities. In evidence path ranking, paths are ranked based on relevance and top-K evidence paths are selected as input for downstream relation extraction models. Alternatively, a causality-guided global reasoning algorithm is also used to filter confusing information and achieve global reasoning to solve cross-document relation extraction~\cite{10153597}. Proposed models for CrossDocRE until now does not consider background knowledge. 

\section{Methodology}

\begin{figure*}[ht]
\includegraphics[width=12cm]{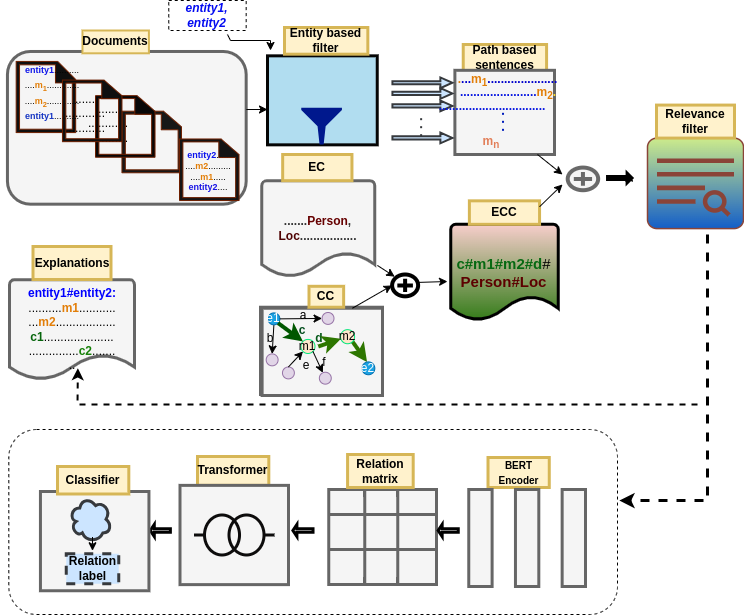}
\centering
   \caption{Architecture diagram of KXDocRE for cross-document relation extraction. Here EC represents entity context, CC represents connecting context and ECC represents both entity and connecting path context.}
    \label{modelcross}
\end{figure*}

\subsection{Problem statement}
For a given entity pair, $\langle e_s,e_o\rangle$ our task is to predict the target relation  $r^c \in \mathcal{R}$ that holds between $e_s$ and $e_o$ within a given corpus of documents $\mathcal{C}_{i=1}^n$, where $\mathcal{R}$ is the relation set and $n$ is total number of documents. If no relation is inferred, it returns the \textit{NA} label. Besides $e_s$ and $e_o$, we introduce the notion of potential bridge entities, which are other mentioned entities in a document that could act as intermediate concepts that help link $e_s$ and $e_o$ via multi-hop connections. The documents provided in CrossDocRE are also annotated with these mentioned entities. We denote $E_i = \{ e^1_i, \ldots, e^m_i \}$ to be the set of $m$ mentioned entities in the $i$-th document, $i=1, \ldots, M$.
CrossDocRE works in two settings, closed and open. In the closed setting, only the related documents are provided to the models, and
the relations are inferred from the provided documents. In the more challenging and realistic open setting, the whole corpus of documents is
provided, and the model needs to efficiently and precisely retrieve related evidence from the corpus. 
We now discuss the architecture of KXDocRE (Figure~\ref{modelcross}) in the subsequent sections.

\subsection{Domain Knowledge}

We consider three forms of domain knowledge knowledge for cross-document relation extraction. 1) The type information of the source and target entity; 2) The connecting path between the source and target entities, extracted using Wikidata; and 3) A combination of 1) and 2). We use the term \emph{context} in the rest of the paper to capture these three forms of knowledge. 

\begin{figure}[t!]
\includegraphics[width=7.5cm]{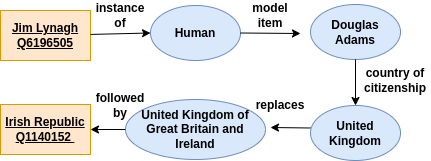}
\centering
   \caption{Context path constructed from Wikidata between Jim Lynagh (source) and Irish Republic (target).}
    \label{context}
\end{figure}

\textbf{Entity type context (EC):} Predicting the relationship between two entities relies heavily on understanding their respective entity types. For instance, if both entities belong to the \emph{Person} category, certain relationships like \emph{has organization}, \emph{has location}, and more are not viable. However, relationships like \emph{child}, \emph{spouse}, or others become possible. Therefore, incorporating the entity type information can assist the model in excluding obvious relationships, thereby improving its performance. Consider the example given in Figure~\ref{context}. The type of the source entity \emph{Jim Lynagh} is \textit{Person}, and the type of the target entity \emph{Irish Republic}, is \textit{Geo Political Entity}, which is used to identify locations, countries, cities, or geopolitical regions. The entity type context for this example is, 
EC$_{\text{\{Q6196505,Q1140152\}}}$=\{Person, GeoPoliticalEntity\}

\textbf{Connecting path context (CC):} The contextual path pertains to an entity pair $\langle e_s,e_o\rangle$. We consider context paths up to $N_h$- h is the hop\footnote{a tunable parameter} distance between the entity pair.
In Figure~\ref{context}, the entity \textit{Jim Lynagh} is five hops away from the entity \textit{Irish Republic}. The four nodes between the entity pair on the path are intermediary entities, and the five edges are the intermediary properties. The contextual path ($CP_{e_i,e_j}$) is formed using the intermediary entities and properties. So, the connecting path context, CC$_{\{\text{Q6196505,Q1140152}\}}$=\{instance of, Human, model item, Douglas Adams, country of citizenship, United Kingdom, replaces, United Kingdom of Great Britain and Ireland, followed by\}.

\textbf{Entity type with connecting path context (ECC):} ECC combines the entity type  and the connecting path context. For Figure~\ref{context}, ECC$_{\{\text{Q6196505,Q1140152}\}}=\text{EC}_{\text{\{Q6196505,Q1140152\}}}$ $\cup$ CC$_{\{\text{Q6196505,Q1140152}\}}$. 

The steps to generate EC, CC, and ECC for a given entity pair are outlined in Algorithm~\ref{algo}. The ContextGeneration function (lines 3-12) computes the EC, CC, and ECC for the given entity pair. CC is constructed in steps of one hop. In lines 14-17, the adjacent edge and node information is retrieved and added to the path. This continues for $N_h$ hops (lines 18-20). The EC of the source and the target entity is obtained using the named entity recognition technique (lines 27-35). 


\begin{algorithm}[ht]
   \small
  \caption{ContextGeneration}
  \label{algo}
      
  \SetKwInOut{KwIn}{Input}
  \SetKwInOut{KwOut}{Output}
  \SetKwProg{Fn}{Function}{:}{}
  \SetKwFunction{path}{ContextGeneration}
  \SetKwFunction{FMain}{ExplorePath}
  \SetKwFunction{FMein}{ExploreEntityType}
  
  \KwIn{ Entity pair (v$_s$,v$_o$),  Number of hops (N$_h$)}
  \KwOut{ Context}
  \DontPrintSemicolon
  
  \textbf{Initialization:} 
  {
   i $\gets$ 1,
     source $\gets$ v$_s$,
     path$_{i-1}$,
     finalcontext,
    finalentitytype,
     adjacent\_node,
     hop\_path$_i$ $\gets$ \textbf{None} }  \\

     {Context} $\gets$ \text{ContextGeneration}(v$_s$, v$_o$, N$_h$)
  
  \Fn{\path{v$_s$, v$_o$, N$_h$}}
  {
    contextpath, entitytype $\gets$  \{\} 
    
    \ForEach{entity pair v$_s$, v$_o$ $\in$ KnowledgeBase}{
   contextpath.append(ExplorePath(v$_s$, v$_o$, N$_h$))\;}
     
        \ForEach{entity pair v$_s$, v$_o$}{
entitytype.append(ExploreEntityType(v$_s$, v$_o$)\;
    }

    finalcontext $\gets$ contextpath  $\cup$ entitytype\;
    \KwRet \{finalcontext\} 
  }
   \Fn{{\FMain}{{(v$_s$, v$_o$, N$_h$)}}}
{
  edge$_i$, adjacent node$_i$ = GetOneHopFromSource(v$_s$) \\
  path$_{i}$ = \{edge$_i$, adjacent node$_i$\}  \\
  i $\gets$ 1 \\
  \While{v$_o$ $\neq$ adjacent node and i $\leq$ N$_h$}{
    path$_{i}$ = path$_{i}$ $\cup$ GetOneHopFromSource(adjacent node)\;
    i $\gets$ i + 1\;
  }
  \If{adjacent node == v$_o$} {
  
  \KwRet \{path$_i$\}\;
}
\Else{
 \KwRet \{\}
}
}
 \Fn{{\FMein}{({v$_s$, v$_o$})}}{
 entity\_type$_{s}$, entity\_type$_o$ $\gets$ \{\} \\
 \If{{EntityTypeExist}(v$_s$)}
{
entity\_type$_s$ $\gets$ \text{GetEntityType}(v$_s$)
}
\If{{EntityTypeExist}(v$_o$)}
{
entity\_type$_o$ $\gets$ {GetEntityType}(v$_o$)
}
finalentitytype $\gets$ entity\_type$_s$ $\cup $ entity\_type$_o$
\KwRet \{finalentitytype\}
}

\end{algorithm}

\subsection{Entity-based filter}
\begin{figure}[ht]
    \centering
    \includegraphics[width=0.35\textwidth]{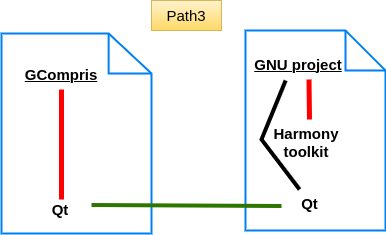}
\caption{An example of a co-occurring graph for path 3 in Figure~\ref{fig:mesh1}.}
    \label{pathdetail}
\end{figure}

In this step, we select sentences that contain meaningful connecting information about the source and target entity and remove irrelevant sentences. The sentences are filtered using the mentioned entity, $e^m$. Similar to the baseline~\cite{wang-etal-2022-entity}, our underlying premise is that a sentence holds significance if it contains either a source or a target entity. Additionally, a sentence is relatively significant if it mentions another entity in conjunction with the entity pair found in both documents. We calculate each entity's score based on three conditions. 1) The source and target entities are in the same sentence as $e^m$ ($\Theta_1$), 2) An entity, $e^o$, co-occurring with $e^m$, also co-occurs with the source and target entities in a different sentence ($\Theta_2$), 3) $e^m$ is part of a text path ($\Theta_3$). Figure~\ref{pathdetail} depicts these conditions using the example from Figure~\ref{fig:mesh1}. The red color represents direct occurrence with the source or target entity in the same sentence, the black line represents indirect co-occurrence, and the green line represents potential co-occurrence. 

For a mentioned entity $e^m$ in each text path $p_i$, the score of each mentioned entity $e^m$ is calculated by the following equation,
$$\small
\begin{aligned}
\text{score}(e^m) &= \lambda S_1(e^m) + \eta S_2(e^m) + \kappa S_3(e^m) \\
S_1(e^m) &= \begin{cases} 1, & \text{if } \Theta_1(e^m) \\ 0, & \text{otherwise} \end{cases} \\
S_2(e^m) &= \begin{cases} \left| \left\{ e^o \mid \Theta_1(e^o) \wedge {I}(e^o) = 1 \right\} \right|, & \text{if } \Theta_2(e^m) \\ 0, & \text{otherwise} \end{cases} \\
S_3(e^m) &= \begin{cases} \left| \left\{ p_i \mid e^m \in \text{E}_i^m \right\} \right|, & \text{if } \Theta_3(e^m) \\ 0, & \text{otherwise} \end{cases}
\end{aligned}
$$


Here, $ \lambda, \eta, \kappa$ are hyper parameters. $I(e^o)=1$ when $e^o$ and $e^m$ are co-occuring in the same sentence, where $e^o\in E_i^m \textbackslash{}
 \{e^m\}$. $S_2$ sums the number of occurrences of $e^o$ and $S_3$ sums the occurrence of mentioned entity in path $p_i$. Next, we calculate the importance score $Imp^s$ of each sentence by aggregating the scores of each mentioned entity,  $Imp^s$=$\sum\limits_{e^m \in E_s^m}$score$\left(e^m\right)$, where  $E_s^m$ is mentioned entity (bridge) in sentence s. The sentences are then ranked based on their importance score, and the top K sentences form the candidate set, $S=\{s_1, s_2, ..., s_K\}$, where K is set to 16 based on the experiment results. We reused the entity based filter from ECRIM model~\cite{wang-etal-2022-entity}.

\subsection{Relevance-based filter}
We aggregate the candidate set sentences from the previous step with the context selected using entity type with the connecting context. 
$S_{total}=S\{s_1, s_2, ..., s_K\}\oplus S_{ECC}$, where $S_{ECC}$ is the string based domain knowledge obtained from ECC. We then apply a relevance-based filter that considers the semantic relevance of the sentence. Here, we assume that if a sentence is semantically similar to another sentence, including the target entity, this sentence should be more informative than other sentences. Our objective is to get the most informative context $I^*$ from the candidate set $S_{total}$ for reasoning about the relation between entities.

\subsection{Encoder}
We mark the start and end of every entity and context using special tokens in sentences. Following the baseline~\cite{yao-etal-2021-codred}, we have used the BERT model~\cite{devlin-etal-2019-bert} to encode the tokens selected from the previous step. Here, $start$ and $end$ are the start and end positions of the j-th mention, $e_j$ is the j-th mentioned entity, and n is the total number of sentences.
\begin{center}
    $e_j=BERT(\{({S_{total}}_{i=1}^n\})_{start}^{end}$
\end{center}

\subsection{Relation matrix}
Besides the mentioned entity, common relations also exist across various text paths. To capture these details from the text path, we used the cross-path entity relation attention module based on Transformer~\cite{10.5555/3295222.3295349}. We collect all entity mentioned representation in a bag and then generate relation representations for entity pairs (e$_u$, e$_v$). Here, $\boldsymbol{E}_r$,  $\boldsymbol{E}_u$, $\boldsymbol{E}_v$ are learnable parameters and e$_u$, e$_v$ are combinations of all entities present, including the mentioned entity.

\begin{center}
    $\boldsymbol{r}_{u, v}=\operatorname{ReLU}\left(\boldsymbol{E}_r\left(\operatorname{ReLU}\left(\boldsymbol{E}_u \boldsymbol{e}_u+\boldsymbol{E}_v \boldsymbol{e}_v\right)\right)\right)$
\end{center}
To model the relation interaction across paths, we build a relation matrix 
$M \in \mathcal{R}^{|E| \times|E| \times d}$, where $E=\bigcup_{i=1}^N E_i$ denotes all the entities in the entity set $E_i$ of text path $p_i$ and $E_i=\left\{e_s, e_o\right\} \cup E_i^m$ and N is total number of entities.
\subsection{Transformer}

For capturing the intra and inter path dependencies, we apply a multi-layer Transformer to perform self attention on the flattened relation matrix $\boldsymbol{\hat {M}} \in \mathbb{R}^{|E|^2 \times d}$.
 \begin{center}
      $\boldsymbol{\hat{M}}^{(t+1|1)} = \text{Transformer}(\boldsymbol{\hat{M}}^{(t)})$

 \end{center}

We obtain the target relation representation $r_{h_{i},t_{i}}$ for each path $p_i$ from the last layer of the Transformer. 

\subsection{Classifier}
After getting the relation representation $r_{h_{i},t_{i}}$  for each text path $p_i$ for each pair of entities, $r_{h_{i},t_{i}}$  is used as a classification feature. We feed these features to the MLP classifier to get the score for each relation. The relation that gets the maximum score is the predicted relation.
\begin{center}
    $\hat{y_i}=\boldsymbol{MLP}(r_{h_{i},t_{i}})$
\end{center}

To obtain the final score for each relation type r, a max pooling operation is applied to each relation label.
\begin{center}
    \( \hat{y}^{(r)}=\operatorname{Max}\left\{\hat{y}_{i}^{(r)}\right\}_{i=1}^{N} \)
\end{center}
After obtaining the scores for all relations, a global threshold $\theta$ is applied to filter out the categories lower than the threshold.
A additional threshold is introduced by baseline paper~\cite{wang-etal-2022-entity} to control which class should be output. The scores of target classes should be greater than threshold and scores of non target class are less than threshold. Formally, for each Bag B, loss is defined as:

\begin{center}
       \( \begin{aligned} \mathcal{L}= & \log \left(e^{\theta}+\sum_{r \in \Omega_{\text {neg }}^{B}} e^{\hat{\boldsymbol{y}}^{(r)}}\right) \\ & +\log \left(e^{-\theta}+\sum_{r \in \Omega_{\text {pos }}^{B}} e^{\hat{\boldsymbol{y}}^{(r)}}\right)\end{aligned} \) 

\end{center}
 
Here, $\hat{y_r}$ is score for relation r , $\theta$ represent threshold and is set to zero,  $\Omega_{\text{pos}}^{B}$ and $\Omega_{\text{neg}}^{B}$  are positive and negative classes between target entity pair.

\subsection{Explanation}
To improve the interpretability of our model, we incorporated an explainable module that can explain the predicted relationship by providing the filtered sentences given to the model. This represents a notable advancement, as contemporary state-of-the-art models cannot often furnish explanations alongside their predictions. In CrossDocRE, the most challenging issue lies in the length of the documents because it affects the amount of noise the model has to handle. So having an explanation module also helps in getting to know how well the model is able to handle noise. Along with that, it also helps in facilitating error analysis.
We retrieve the tokens that are fed to the model ($I^*$), converting them into strings. This way, we get the exact token data that was used to make the relation prediction. This process enables us to obtain the precise token data, which drives predictions. 


\section{Experimental setup}

\textbf{Metrics.} We used F1 and area under curve (AUC) scores following the baseline~\cite{yao-etal-2021-codred} for a fair comparison. For the open setting, we retrieve the top k document paths from the Wikipedia corpus\footnote{We follow the baseline paper and set k to 16~\cite{yao-etal-2021-codred}} and use the models trained in the closed setting to predict a relation. These paths are scored using three conditions. 1) Entity count: the number of occurrences of the source entity in the document containing the source and likewise with the target entity. 2) Shared entity: the number of shared entities that  appear in source and target documents 3) TF-IDF: The TF-IDF similarity between the two documents. Similar to baseline, we selected the top 16 paths with the highest scores. Hyper-parameters used in our model are shown in Table~\ref{crosshyper}.

\begin{table}[ht]
\centering
\scalebox{0.8}{
 \begin{tabular}{cc} 
 \hline
  \textbf{Hyper parameters} &\textbf{Value } \\
  \hline
    Learning rate & 3e-5 \\
    Embedding dimesion & 768 \\
    Encoder layers & 3\\
    $\gamma, \eta, \kappa $& 0.1,0.01, 0.001 \\
    Optimizer & AdamW \\
   \hline

 \hline

 \end{tabular}
 }
 
 \caption{Hyper-parameters setting}
 \label{crosshyper}

\end{table}
\textbf{Baseline model for comparison.} We  used all the baseline models available in CrossDocRE for comparison. 1) CodRED~\cite{yao-etal-2021-codred} is used, which extracts text snippets surrounding the two entities in the document as input and feeds it into a BERT-based model
2) ECRIM~\cite{wang-etal-2022-entity}; an entity-based document filter is constructed and then fed into a BERT-based model.
3) MR.COD~\cite{Lu2022MultihopER} is a multi-hop evidence retrieval method based on evidence path mining
and ranking. 4) LGCR~\cite{10153597} discusses local to global reasoning method  which enables efficient distinguishing, filtering and global reasoning on complex information from a causal perspective.

We conducted our evaluation in response to the following research questions. \textbf{RQ1}: What is the effectiveness of KXDocRE in combining context knowledge with reasoning in solving CrossDocRE? \textbf{RQ2}: Does the explanation generated by our approach provide sufficient grounds to support the inferred relation?

\textbf{Dataset.}
We used the CodRED dataset for evaluation. CodRED contains 11,971 entities and 276 relation types in this dataset. The details of CoDRED dataset are given in Table~\ref{datasetcodred}. Bags represent relational facts of dataset, 2733 are positive labeled facts and 16,668 are labeled as NA in training set.
\begin{table}[ht]
\centering

\scalebox{0.7}{

\begin{tabular}{ccccc}

\hline
&  & {Train} & {Dev} & {Test} \\
\hline
\multirow{2}{4em}{Bags} & Pos & 2733  & 1010 &  1012 \\ 
& NA& 16,668 & 4558 & 4523 \\ 
\hline
Text paths& - & 129,548 & 40,740 & 40,524 \\ 
\hline
\end{tabular}
}

\caption{Statistics of CoDRED dataset}
\label{datasetcodred}
\end{table}

\section{Results}
Our model works in the open and closed settings of CrossDocRE. We evaluated our model with three variations: using entity context (EC), connecting context (CC), and both (ECC). We used an NVIDIA A100-SXM4 tensor core GPU with 40GB of memory on Linux 5.4.0-125 with Python version 3.8.5. The results for closed and open settings are available in Table~\ref{openresults}. The results on the test set were obtained from CodaLab\footnote{\url{https://codalab.lisn.upsaclay.fr/competitions/3770}}. 
In contrast to a closed setting, performance declines in an open setting due to the retrieval of paths, not all of which significantly contribute to reasoning.
Compared to the baseline model, KXDocRE improves by $\approx$ 3\% in the F1 score (closed setting) and $\approx$ 4\% in the F1 score (open setting). In all the settings of KXDocRE, it can reason over text better than baseline. Our findings demonstrated the significant role domain knowledge can play in the process of reasoning. Hence, we answer \textbf{RQ1}. 
In addition to BERT, we did attempt using RoBERTa~\footnote{https://huggingface.co/docs/transformers/en/model\_doc/roberta} and GPT2~\footnote{https://huggingface.co/docs/transformers/en/model\_doc/gpt2} in our evaluations. However, there was no improvement in the F1 score. Hence, we decided to report BERT scores.

\begin{table*}[ht]
        \centering
          \begin{minipage}[t]{0.45\textwidth}

\scalebox{0.6}{
\begin{tabular}{cccccc}
\hline
\multicolumn{6}{c}{\textbf{Closed setting}} \\
\hline
\textbf{Baseline model}  & \textbf{PLM/GNN} & 
\multicolumn{2}{c}{\textbf{Dev}} & \multicolumn{2}{c}{\textbf{Test}} \\
\cline{1-6} 
{}  & {} & \textbf{F1} & \textbf{AUC} & \textbf{F1} & \textbf{AUC} \\

\hline
{CoDRED} & BERT & 51.26 & 47.94 & 51.02 & 47.46 \\

 {ECRIM}& BERT & 61.12 & 60.91 & 62.48 & 60.67 \\
 {MR.COD}& BERT & 61.2 & 59.22 & 62.53 & 61.68 \\
 {LGCR}& BERT & 61.67 & 63.17 & 61.08 & 60.75 \\
  {LGCR}& RoBERTa & 63.18 & 64.76 & 63.79 & 63.03 \\

  {\textbf{KXDocRE$_{EC}$}}& \textbf{BERT} & \textbf{63.57} & \textbf{62.8}& \textbf{65.3} & \textbf{64.45} \\
 
  {\textbf{KXDocRE$_{CC}$}}& \textbf{BERT} & \textbf{64} & \textbf{63.7}& \textbf{65.8} & \textbf{64.90} \\
{\textbf{KXDocRE$_{ECC}$}}& \textbf{BERT} & \textbf{64.97} & \textbf{64.30}& \textbf{66.3} & \textbf{65.55} \\
\hline
 
\end{tabular}
}
  \end{minipage}
  \hfill
        \centering
          \begin{minipage}[t]{0.45\textwidth}

        \scalebox{0.6}{
        \begin{tabular}{cccccc}
        \hline
\multicolumn{6}{c}{\textbf{Open setting}} \\

\hline
\textbf{Baseline model}  & \textbf{PLM/GNN} & 
\multicolumn{2}{c}{\textbf{Dev}} & \multicolumn{2}{c}{\textbf{Test}} \\
\cline{1-6} 
{}  & {} & \textbf{F1} & \textbf{AUC} & \textbf{F1} & \textbf{AUC} \\

\hline
{CoDRED} & BERT & 47.23 & 40.86 & 45.06 & 39.05 \\
 {MR.COD}& BERT & 53.06 & 51.00 & 57.88 & 53.30 \\
 {LGCR}& BERT & 52.96 & 51.48 & 53.45 & 50.15 \\
  {LGCR}& RoBERTa & 55.15 & 52.36 & 55.37 & 49.05 \\

  {\textbf{KXDocRE$_{EC}$}}& \textbf{BERT} & \textbf{55.5} & \textbf{54.3}& \textbf{56.15} & \textbf{50.11} \\

  {\textbf{KXDocRE$_{CC}$}}& \textbf{BERT} & \textbf{55.9} & \textbf{54.8}& \textbf{57.12} & \textbf{50.6} \\
  {\textbf{KXDocRE$_{ECC}$}}& \textbf{BERT} & \textbf{56.7} & \textbf{55.2}& \textbf{57.93} & \textbf{57.12} \\
 
\hline
\end{tabular} }

  \end{minipage}
  \caption{Results on CodRED dataset for closed and open setting}
            \label{openresults}
\end{table*}

\subsection{Ablation Study}
\begin{figure}[ht]

\includegraphics[scale=0.35]{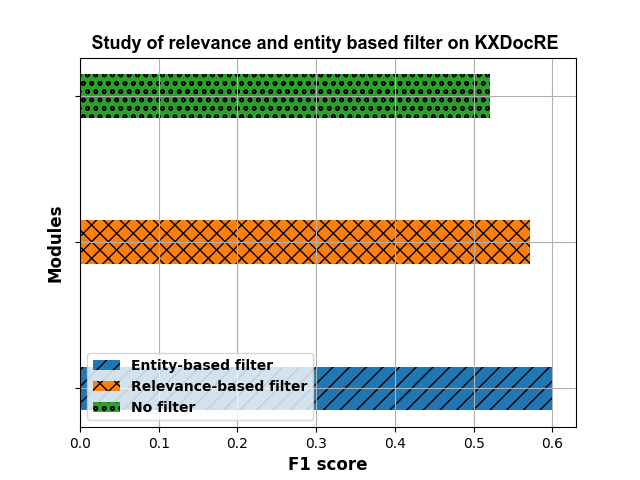}
\caption{Study of relevance and entity based filter on KXDocRE}
\label{ablationcr}

\end{figure}

\textbf{Effectiveness of relevance and entity-based filter:} We studied the impact of relevance and entity-based filters on the performance of KXDocRE. Figure~\ref{ablationcr} shows the F1 score obtained using the Dev dataset. After removing the relevance filter, the performance dropped by 7.7\% and if the entity-based filter is removed, the performance dropped by 4.9\%. After removing both filters, the F1 score drops significantly by 12\%. This indicates that the two filters play an important role in KXDocRE. \\

\begin{figure}[ht]
    \centering
    \includegraphics[scale=0.35]{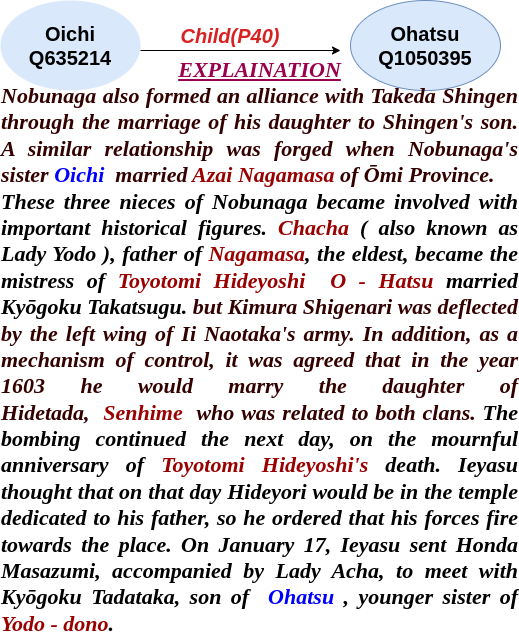}
\caption{Explanation generated using KXDocRE. The source and target entities are blue in color.  Intermediate entities are in red, and aggregated context is green in color}
    \label{explain}
\end{figure}

\begin{figure}[htb]
    \centering
    \includegraphics[scale=0.35]{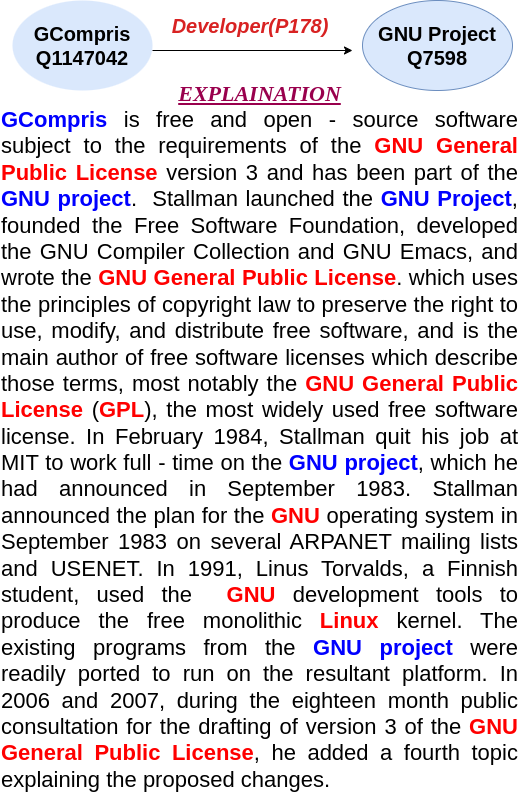}
\caption{Explanation generated from KXDocRE for example discussed in Figure~\ref{fig:mesh1}}
    \label{explaination}
\end{figure}

\textbf{Effectiveness of explanations:} 
In cross-document setting, predicting the relationship between entity pairs involves considering multiple text paths (averaging 3646) found across various documents. The explanation module aids in surfacing the most relevant paths that influence the prediction. This module is expected to help us to understand the prediction using a single paragraph containing the relevant paths, rather than scrutinizing each document individually.
We discuss the explanation module using a case study (Figure~\ref{explain}). The labeled relation between Oichi (Q635214) and Ohatsu (Q1050395) is \emph{Child (P40)}. The explanation text provides text paths between Oichu and Ohatsu via Azai, Toyotomi and Yodo-dono (Oichi is the spouse of Azai Nagamasa, who is the child of Yodo-dono, who is the sibling of Ohatsu); hence, Oichi is the child of Ohatshu. Therefore, the explanation text provides enough evidence along with the context to reason over the data to understand the predicted relationship. Hence, we answer \textbf{RQ2}.
We depict another explanation Fig ~\ref{explaination} using KXDocRE for example discussed in Fig~\ref{fig:mesh1}.

\begin{figure}[htb]
    \centering
    \includegraphics[scale=0.30]{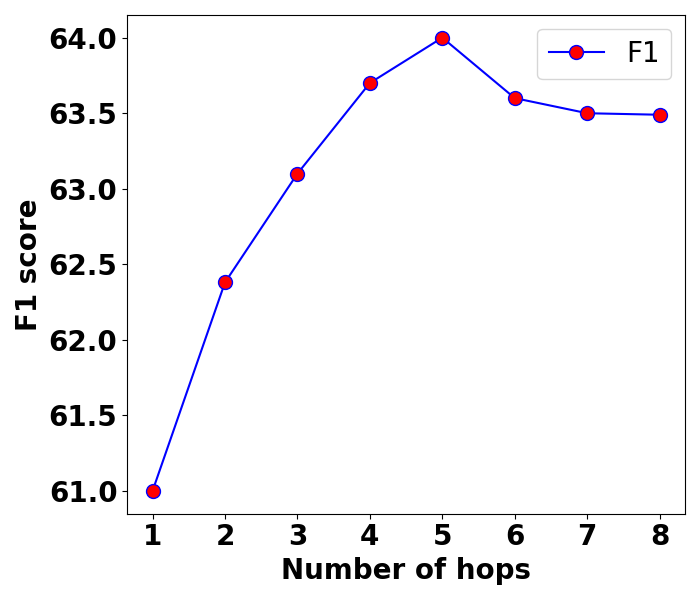}
\caption{Effectiveness of hops in KXDocRE}
    \label{hops}
\end{figure}

\textbf{Effectiveness of number of hops on KXDocRE:}
We also studied the impact of the number of hops considered for extracting context in a CC setting on the F1 score (Figure~\ref{hops}). The F1 score increases with the number of hops until a point, and after that, the F1 score starts decreasing and saturates. Increasing the hops do not add much new and relevant information after a certain point. Due to this reason, we considered the number of hops ($N_h$) up to 5.

\textbf{Error analysis:}
We studied the successful and failed cases of KXDocRE based on the domain knowledge of the entity pair. From Table~\ref{erroranalysis}, we can say that the likelihood of a correct prediction is higher if context is available for the given entity pair. \\

\begin{table}[htb]
\centering
\scalebox{0.58}{
 \begin{tabular}{cccccc} 
 \hline
 & \textbf{\#  Dev} &\textbf{\# Correct } 
   &\textbf{\# Incorrect }  & \textbf{Correct}\\  
   & \textbf{entity pair}  & \textbf{prediction} & \textbf{prediction} & \textbf{prediction\%} \\
   \hline

 KXDocRE & 5567 &  4523 & 1044 & 81.2 \\
 (Total entity pair) & & & & \\
\hline
 
 KXDocRE$_{EC}$ & 952 &  894 & 58 & 93.9 \\
 
 KXDocRE$_{CC}$ & 921 &  890 & 31 & 96.6 \\

 \hline

 \end{tabular}
 }
 \caption{Impact of context in successful cases on Dev dataset }
 \label{erroranalysis}

\end{table}

\textbf{Complexity analysis:} In the CodRED dataset, the text path consists of 129,548 instances in the training set, 40,740 instances in the development set, and 40,524 instances in the test set. The average time of a single epoch's execution in the baseline and our model is given in Table~\ref{complexity}. The execution time for KXDocRE is longer (linear increment) due to the addition of context in the module. The time taken to create domain knowledge is available in Table~\ref{complexity1}. 

\begin{table}[H]
\centering
\scalebox{0.6}{
\begin{tabular}{ccccc}
\hline
  \textbf{Model} & \textbf{Total Time} & \textbf{Total} & \textbf{Average time (in seconds)} \\
& \textbf{(in hours)} & \textbf{(\#epochs)} & \textbf{per epoch}\\
\hline
 ECRIM (baseline) &  76 & 10 & 7.6\\ 
KXDocRE(ours) & 96.8 &  10 & 9.6 \\
\hline

\end{tabular}
}
 \caption{Complexity analysis of KXDocRE as compared to baseline.}
 \label{complexity}

\centering
\scalebox{0.58}{
 \begin{tabular}{cc} 
 \hline
 \textbf{Module} & \textbf{Average time (in sec)}  \\ [0.5ex] 
 \hline
 Entity context  & 0.001 \\ 
 Context path (1-hop) & 0.11  \\
 Context path (2-hop)  & 0.21   \\
 Context path (3-hop)  & 0.38  \\ [1ex] 
 \hline
 \end{tabular}
 }
 \caption{Average time taken to create the context in KXDocRE.}
 \label{complexity1}

\end{table}

\textbf{Case study:} We discuss two successful cases and one failed case of KXDocRE and compared them with the baseline model (Table~\ref{crpsscasestudy}). \textbf{Case1:} To identify the relation between \emph{Dreamlover (Q909801)} and \emph{If it's Over (Q1095958)} from Documents 1 and 2, we use domain knowledge of both entities. EC does not exist for the given entity pair, hence we use CC. CC for both entities is \{\emph{part of, Emotions, tracklist, If It's Over, followed by}\}. Adding CC helps KXDocRE to predict the correct relation compared to our baseline model, ECRIM.\textbf{ Case2:} In second case study, for entity pairs \emph{Airbus A320neo family (Q6488)} and \emph{Airbus (Q67)}, we add the ECC context as \{\emph{owned by, ORG, ORG}\}, which helps KXDocRE help in predicting the relation.
\textbf{Case3:} We studied a failed case of KXDocRE for entity pairs \emph{Adium (Q58058)} and \emph{x86\_64 (Q272629)}. CC for this entity pair is \{\emph{instance of, free software, subclass of, software, model item, Mozilla Firefox model item}\}. This context does not contribute significantly in predicting the relationship.

\begin{table}[ht]
\centering
\small 
\noindent
\scalebox{0.67}{
\begin{tabular}{@{} p{1.5\linewidth}}
\hline

 \textbf{Case1}: \\
 \hline
\colorbox{green}{Document 1}: \textcolor{blue}{Dreamlover} is a song by American singer \textcolor{red}{Mariah Carey} ....\textcolor{blue}{Dreamlover} marked a more pronounced attempt on \textcolor{red}{Carey's} part to incorporate .... arey began to alter her songwriting style and genre choices, most notably in \textcolor{blue}{Dreamlover}. \textcolor{blue}{Dreamlover} an Vision of Love \textcolor{red}{Carey's} best, calling them the original hits............................ \\

\colorbox{green}{Document 2}:
If It's Over \textcolor{blue}{If It's Over} is a song written by American singers and songwriters \textcolor{red}{Mariah Carey} and .....\textcolor{blue}{if it's over}, \textcolor{red}{let me go} Several months after the release of \textcolor{red}{Emotions} \textcolor{red}{Carey} performed the song...\textcolor{red}{If It's Over} is a downtempo \textcolor{red}{ballad}, which incorporates several genres........... 

 \textbf{Correct answer}:  \textcolor{violet}{\textbf{followed by}} \\
 \textbf{CodRED}: N/A \\
 \textbf{ECRIM}: N/A \\
 \textbf{KXDocRE$_{ECC}$}: \textcolor{purple}{followed by}  \\ 
 \hline
  \textbf{Case2}: \\
\hline

\colorbox{green}{Document 1}: It is the core development area of of \textcolor{red}{Bandai Namco} group. ......main video game branch of \textcolor{blue}{Bandai Namco Holdings}. In Feb 2005, primariliy set in fictional \textcolor{red}{Japanese} city of ......, in association with the \textcolor{red}{Japanese} government, suppressed information can be...................... \\
\colorbox{green}{Document 2}: \textcolor{blue}{Bandai Visual}, Bandai Entertainment, Dentsu, Nippon...........and original \textcolor{red}{Japanese} (one late-night screening).......It was released on 4 March 2004 in \textcolor{red}{Japan} and 8 November 2004 in \textcolor{red}{North America}.....The second volume Ghost in the shell.............in \textcolor{red}{Japan} and on 26 September 2006......

 \textbf{Correct answer}: \textcolor{violet}{\textbf{parent organization}} \\
 \textbf{CodRED}: N/A \\
  \textbf{ECRIM}: N/A \\
 \textbf{KXDocRE$_{ECC}$}:\textcolor{purple}{parent organization} \\
 \hline
  \textbf{Case3}: \\
\hline

\colorbox{green}{Document 1}: \textcolor{blue}{Adium} is a free and open source instant messaging client for \textcolor{red}{macOS} that supports... including \textcolor{red}{Windows Live Messenger}...It is written using \textcolor{red}{macOS}...under the \textcolor{red}{GNU}....\textcolor{blue}{Adium} makes use of a plugin \textcolor{red}{architecture}...\\
\colorbox{green}{Document 2}: In computer \textcolor{red}{architecture}, \textcolor{red}{64-bit} integers, \textcolor{red}{memory}.....ALU \textcolor{red}{architectures} are those that are based on.........\textcolor{red}{AMD} released its first \textcolor{blue}{x86-64}...\textcolor{red}{Java} program can run on a 32- or ... \\
 \textbf{Correct answer}: \textcolor{violet}{\textbf{N/A}} \\
 \textbf{CodRED}: N/A \\
  \textbf{ECRIM}: N/A \\
 \textbf{KXDocRE$_{ECC}$}:\textcolor{purple}{N/A} \\
 \hline


\hline
\end{tabular} 
}
\caption{Case study}
\label{crpsscasestudy}
\end{table}

\section{Conclusion}
We introduce a novel model called KXDocRE with three settings (EC, CC, ECC) that incorporate domain knowledge in CrossDocRE. Our results suggest that model performance is improved by integrating diverse domain knowledge. Also, we provide an explanation text for each prediction that makes our model interpretable. As a future work, researchers can create a versatile model that can work on diverse document sets, accumulating domain knowledge.

\section{Limitations}

Our work has limitations in terms of creating a connecting context. A connecting context will only be created if there is some path between two entities in Wikidata. We plan to overcome this limitation by incorporating other external knowledge bases for creating context. With the increase in the number of text paths and mentioned entities, the GPU memory consumption increases, and the speed decreases.

\section{Acknowledgement}
We express our sincere gratitude to the Infosys Centre for Artificial Intelligence (CAI) at IIIT-Delhi for their support. RK's effort has been supported by the U.S.~National Library of Medicine (through grant R01LM013240)

\bibliography{latex/references}

\appendix


\end{document}